\documentclass[letterpaper]{article} 
\usepackage[preprint]{aaai2027}  
\usepackage[hyphens]{url}  
\usepackage{graphicx} 
\usepackage{natbib}  
\usepackage{caption} 
\usepackage{algorithm}
\usepackage{algorithmic}
\usepackage{amsmath}
\usepackage{amssymb}
\usepackage{newfloat}
\usepackage{listings}
\DeclareCaptionStyle{ruled}{labelfont=normalfont,labelsep=colon,strut=off} 
\floatstyle{ruled}
\newfloat{listing}{tb}{lst}{}
\floatname{listing}{Listing}

\usepackage{booktabs}

\newcommand{\method}{\textsc{HyMeS}}

\title{Skills in Weights, Memory in Code: Hybrid Learning for Memory-Dependent Robot Manipulation}

\author{
Yunhao Zhao\textsuperscript{\rm 1*}, \quad
Zhenyang Ni\textsuperscript{\rm 1*}, \quad
Haoyang Chen\textsuperscript{\rm 2}\\
Ruohan Zhang\textsuperscript{\rm 1,3}, \quad
Qi Zhu\textsuperscript{\rm 1}
}

\affiliations{
\textsuperscript{\rm 1}Northwestern University \quad
\textsuperscript{\rm 2}University of Minnesota \quad
\textsuperscript{\rm 3}Stanford University\\[0.3em]
{\scriptsize\texttt{yunhaozhao2028@u.northwestern.edu}} \quad
{\scriptsize\texttt{zhenyangni2030@u.northwestern.edu}} \quad
{\scriptsize\texttt{chen9861@umn.edu}}\\
{\scriptsize\texttt{zharu@stanford.edu}} \quad
{\scriptsize\texttt{qzhu@northwestern.edu}}\\[0.3em]
\textsuperscript{*}Equal contribution
}

\begin{document}

\maketitle

\begin{abstract}
Modern vision-language-action (VLA) policies have acquired broad manipulation skills, but typically generate each action chunk from the current observation or a short fixed-length history. However, real-world manipulation is often non-Markovian, requiring robots to retain and reason over task-relevant information from long-horizon interaction histories to determine the next action. To address this challenge, we propose \textsc{HyMeS}, a \textbf{Hy}brid learning framework that leverages the reasoning and \textbf{Me}mory-management capabilities of coding agents to \textbf{S}teer a Markovian VLA for memory-dependent manipulation. Specifically, \textsc{HyMeS} learns low-level motor skills through gradient-based imitation learning, while a coding agent acquires high-level memory-management strategies through heuristic learning by iteratively updating an executable heuristic system from rollout feedback. Furthermore, we close the loop between steering and execution through multimodal stage-completion verification, which updates memory using proprioceptive signals and multi-frame VLM judgments. Compared with end-to-end memory-augmented VLAs, HyMeS requires demonstrations only for reusable motor skills rather than for every history-dependent task configuration, enabling data-efficient compositional generalization. On RoboMemArena, HyMeS improves mean cumulative success from 52.5\% to 66.2\% and mean task success from 41.3\% to 60.1\% over $\pi_{0.5}$, while outperforming PrediMem by 4.5 points in cumulative success and 14.5 points in task success.

\end{abstract}


\section{Introduction}

\begin{figure}[!t]
\centering
\includegraphics[width=1.0\columnwidth]{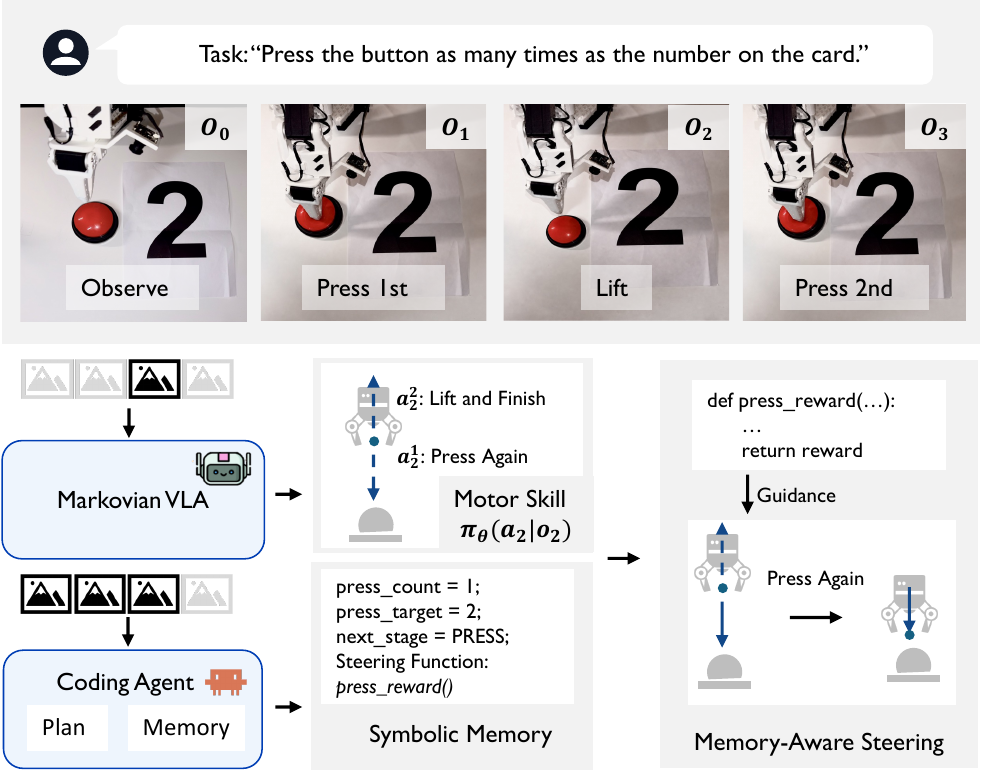}
\caption{\textbf{Memory as code resolves history-dependent action ambiguity.}
\emph{Top:} the required number of presses is specified by a card observed
at the start of the episode. \emph{Bottom:} at $o_2$ the first press is
complete and the arm has lifted, so the current observation alone supports
two incompatible action modes: press again ($\mathbf a_2^{1}$) or lift and
stop ($\mathbf a_2^{2}$). The Markovian VLA
 conditions on $o_2$ only and cannot
choose between them. \method{} instead passes the interaction history to a
coding agent that maintains symbolic memory in executable form, and selects the corresponding
steering function. The gradient of that function guides action generation
toward pressing again, using the same policy weights with no
weight update.}
\label{fig:teaser}
\end{figure}

Vision--language--action (VLA) models have made rapid progress toward general-purpose robot control. Recent models such as $\pi_{0.5}$ generalize across diverse objects, scenes, and language instructions \citep{pi05}, while $\pi_{0.6}$ further extends these capabilities to fine-grained and dexterous manipulation \citep{pi06}. A single policy can now acquire a broad repertoire of motor skills and achieve high success rates across many manipulation tasks. These advances suggest that low-level motor competence is increasingly well served by modern VLAs.

Despite this progress, most VLAs remain Markovian by design: they generate each action chunk from only the current observation or a short fixed-length context, whereas real-world robot manipulation is often non-Markovian and requires reasoning over task-relevant information from long-horizon interaction histories. For example, although pressing a button is a simple motor skill, a Markovian VLA cannot reliably press multiple buttons the required numbers of times when the counts must be recalled from earlier observations. Existing memory-augmented VLAs address this limitation by extending the observation context or adding learned memory modules, then training the history-conditioned policy end-to-end from robot demonstrations \citep{memoryvla,eventvla,memorywam}. However, this entangles memory acquisition with motor learning, forcing expensive robot demonstrations to cover combinatorial memory configurations even when no new physical skill is involved.

To overcome this limitation, we propose \method{}, a \textbf{Hy}brid learning framework that leverages the reasoning and \textbf{Me}mory-management capabilities of a coding agent to \textbf{S}teer a Markovian VLA. The core idea is to separate motor-skill learning from memory-strategy learning. The VLA acquires low-level motor skills in weight space through gradient-based imitation learning, while the coding agent acquires high-level memory-management strategies in code space through heuristic learning \citep{weng}. The coding agent iteratively updates an executable heuristic system from rollout feedback. During execution, this system maintains symbolic task memory and translates it into memory-conditioned constraints that steer the VLA's flow-matching action generation. We further introduce a multimodal stage-completion verification mechanism that closes the loop between steering and execution by updating memory from proprioceptive signals and multi-frame VLM judgments.

Compared with end-to-end memory-augmented VLAs, \method{} requires demonstrations only for reusable motor skills rather than for every history-dependent task configuration, enabling data-efficient compositional generalization. We evaluate \method{} on four memory-dependent task categories in \textsc{RoboMemArena} against the same-weight $\pi_{0.5}$ policy and memory-augmented PrediMem. \method{} improves Cumulative Success Rate (CSR) from 52.5\% to 66.2\% and Task Success Rate (TSR) from 41.3\% to 60.1\% over $\pi_{0.5}$, while exceeding PrediMem by 4.5 CSR and 14.5 TSR points. On the SO-101 robot, it raises TSR from 25.7\% to 57.1\%. Ablations show that rollout experience improves CSR and TSR by 8.5 and 18.4 points, while combining proprioceptive and visual signals enables more reliable stage transitions than either signal alone.

Our contributions in this work include:
\begin{itemize}
    \item We formulate memory-dependent manipulation as a \textbf{hybrid learning} problem that separates motor-skill learning from memory-strategy learning. Motor skills are learned in weight space from expert demonstrations, while memory-management strategies are acquired in code space from rollout feedback through heuristic learning.
    
    \item We introduce \method{}, which uses executable symbolic memory to steer a Markovian VLA through memory-conditioned denoising constraints. We further close the loop between steering and execution with multimodal stage-completion verification that updates memory from proprioceptive signals and multi-frame VLM judgments.
    
    \item On \textsc{RoboMemArena}, \method{} improves CSR from 52.5\% to 66.2\% and TSR from 41.3\% to 60.1\% over $\pi_{0.5}$, while exceeding PrediMem by 4.5 and 14.5 points, respectively.
\end{itemize}

\FloatBarrier
\begin{figure*}[t]
    \centering
    \makebox[\textwidth][c]{%
        \includegraphics[width=1.08\textwidth]{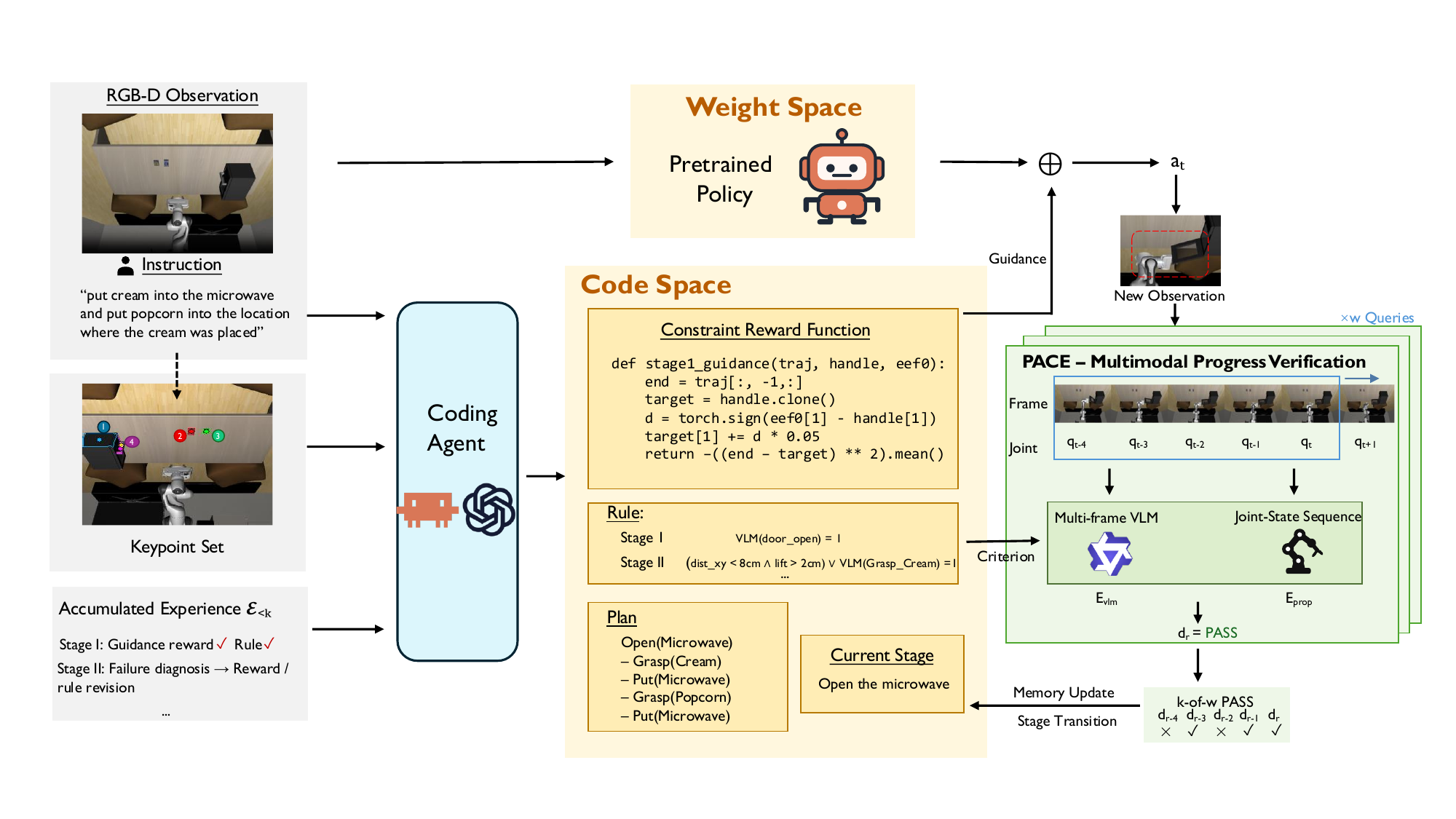}%
    }
    \caption{\textbf{Overview of the \method{} within-episode execution loop.}
    The instruction and the current observation are passed to the pretrained policy in \emph{weight space}, whose weights stay fixed throughout. At initialization and after each verified stage transition, SAM and DINOv2 extract task-relevant keypoints; the coding agent (claude code / codex) combines them with the observation and the program $\mathcal P^{(n)}$ refined over prior rollouts to instantiate a \emph{code-space} memory holding an executable plan, the current stage, its verification rules, and a stage-specific constraint reward. The gradient of that reward is injected into action generation as guidance, so the executed action chunk $\mathbf{a}_t$ follows the current memory state without any weight update. PACE verifies progress using synchronized multi-frame visual and joint-state evidence, advancing the stage only when at least \(k\) of the most recent \(w\) completion queries pass.}
    \label{fig:system}
\end{figure*}

\section{Related Work}

\textbf{Memory-augmented VLAs.} Modern VLA models map the current observation---or a short, fixed-length window---to actions \citep{rt1,rt2,openvla,diffusionpolicy,pi0,pi05}, and are thus Markovian by construction. The dominant remedy for memory-dependent tasks bakes the missing memory into the action model and learns it end-to-end: by pretraining history representations \citep{aem}, retrieving perceptual--cognitive memories \citep{memoryvla,memoryvlapp}, recurring over memory tokens \citep{muvla}, recoding adaptive working memory \citep{memoact,pam}, or storing event-triggered, hierarchical, and full-history states \citep{eventvla,memorywam,dimwam,himemwam,chronos}. Because the memory logic is combinatorial and discrete, learning it in weight space demands demonstrations that cover the combinatorial space, is prone to forgetting, and yields latent buffers that are hard to inspect or edit. \method{} instead trains the action model only for motor skills, keeping it Markovian, and acquires the memory logic in code space---without demonstrations that cover the memory combinatorics and without any memory-aware weight updates.

\smallskip
\noindent
\textbf{Steering VLAs at inference time.} A complementary line keeps a pretrained generative policy frozen and biases its action sampling at test time, injecting guidance from human interactions \citep{itps}, learned value functions \citep{vgps}, goal-conditioned dynamics models \citep{dynaguide}, success predictors \citep{ppguide}, tactile feasibility \citep{touchguide}, or VLM-synthesized stage rewards \citep{vls}. These methods preserve the base policy as a skill prior, but their guidance is reactive---computed from the current observation or an external signal---so none maintains the episode-level memory that memory-dependent tasks demand. Moreover, information flows in only one direction, with nothing read back from the policy. \method{} adopts the steering interface of VLS but conditions the injected constraint on an explicit symbolic memory, and reads execution feedback (proprioceptive and visual) back for event detection, making the coupling bidirectional.

\smallskip
\noindent
\textbf{Coding agents for robotics.} LLMs that write executable code have long served robotics as policies and reward designers: Code as Policies and VoxPoser compose control APIs and value maps from language \citep{cap,voxposer}, and Eureka evolves reward code for skill learning \citep{eureka}. Recent agentic systems close the loop with environment feedback, iteratively rewriting controllers or policy repositories from rollout outcomes \citep{faea,alrm,aor,rho}, while orchestration frameworks let a VLM agent plan over, invoke, and monitor low-level policies as tools \citep{volo,goal2skill}. Most closely related, Harness VLA uses task-specific traces and global failure rules to orchestrate a frozen VLA as a retryable contact-rich primitive \citep{zhang2026harnessvlasteeringfrozen}. Its memory operates at primitive granularity, whereas \method{} couples executable memory directly to the VLA's denoising process to steer individual actions. In all of these, however, the agent's code either \emph{replaces} the low-level policy or \emph{commands} it at subtask granularity, leaving action generation itself untouched. In \method{}, the coding agent instead maintains memory as executable task state and learns its update rules in code space from rollout traces \citep{weng}, coupling this memory to the inside of a Markovian VLA's denoising loop to steer action generation step by step.

\section{Problem Formulation}

We consider robot manipulation under a language instruction
$\ell$. At control step $t$, the robot receives an observation
$o_t$, consisting of RGB images and proprioception, and executes
an action chunk $\mathbf a_t$. We denote the interaction history
by $h_t=(o_1,\mathbf a_1,\ldots,o_{t-1},\mathbf a_{t-1},o_t)$.
A task is \emph{memory-dependent} when the expert policy depends
on the history only through a compact task state
\begin{equation}
z_t=\phi(h_t),
\qquad
\pi^\ast(\mathbf a_t\mid h_t,\ell)
=
\pi^\ast(\mathbf a_t\mid o_t,z_t,\ell),
\label{eq:task_state}
\end{equation}
where $z_t$ (e.g., a cue observed once and then removed, or the
number of completed repetitions) is not recoverable from $o_t$
alone. The observation is thus \emph{aliased}: histories with
different task states can yield the same $o_t$ while demanding
incompatible actions (Fig.~\ref{fig:teaser}), so a Markovian
policy $\pi_\theta(\mathbf a_t\mid o_t,\ell)$ must average over
the induced action modes and cannot match $\pi^\ast$.

We assume an expert demonstration set
$\mathcal D=\{(o_i,\mathbf a_i,\ell_i)\}_{i=1}^{N}$ that covers
the motor skills required by the task but does not enumerate the
combinatorial space of task states $z$. During method
development, the robot can additionally collect interaction
rollouts containing execution traces and stage-wise task
outcomes, but no expert action labels. Our goal is to learn a
Markovian motor policy from $\mathcal D$ and, from rollout
feedback, an external memory mechanism that maintains an
executable estimate $s_t$ of $z_t$ and injects it into action
generation, yielding history-consistent actions without
demonstrations covering each task state.

\section{Method}

\label{sec:overview}

We propose \method{}, a hybrid learning framework comprising
a flow-matching VLA and an executable heuristic program. The
two components acquire complementary capabilities through
different learning mechanisms. The VLA learns motor skills in
weight space through gradient-based imitation learning. The
heuristic program learns memory-management strategies in code
space through rollout-driven revisions by a coding agent. After
the VLA has been fine-tuned, its parameters remain fixed throughout
heuristic development and evaluation.

The learned program
$\mathcal P=(\mathcal C,\mathcal V,\mathcal U)$ contains
constraint-selection rules $\mathcal C$, event-verification
rules $\mathcal V$, and memory-update rules $\mathcal U$.
Let $\pi_{\theta^\star}$ denote the pretrained VLA after
task-specific motor-skill fine-tuning, and let $s_t$ be the
symbolic memory state maintained by $\mathcal P$ at control
step $t$---the executable estimate of the latent task state
$z_t$ in Eq.~\ref{eq:task_state}.
t evaluation, $\mathcal P$ is frozen: evaluation rollouts no longer
revise its heuristic experience or constraint-generation strategy.
The coding agent still applies $\mathcal P$ to instantiate
context-dependent stage rules and the initial memory
$s_1=\operatorname{Init}_{\mathcal P}(\ell,o_1)$,
building the stage plan from the instruction and the initial
observation (Fig.~\ref{fig:system}), and then interacts with
the Markovian VLA through the following closed-loop process:
\begin{align}
R_t &=
\mathcal C(s_t,o_t,\ell),
\label{eq:overview_constraint}\\
\mathbf a_t &=
\operatorname{Steer}
\bigl(\pi_{\theta^\star},R_t,o_t,\ell\bigr),
\label{eq:overview_steer}\\
e_t &=
\mathcal V
\bigl(o_{\mathcal W_t},\mathbf a_{\mathcal W_t};s_t\bigr),
\label{eq:overview_event}\\
s_{t+1} &=
\mathcal U(s_t,e_t,o_t),
\label{eq:overview_update}
\end{align}
where $R_t$ is the differentiable constraint selected for the
active stage (Eq.~\ref{eq:reach_constraint}),
$\operatorname{Steer}(\cdot)$ denotes constraint-guided action
generation (Eq.~\ref{eq:guidance}), and $e_t\in\{0,1\}$ is the
stage-progress event verified by $\mathcal V$ from the
observations and executed action chunks in a sliding
verification window $\mathcal W_t$, instantiated by the
multimodal vote of Eq.~\ref{eq:completion_vote}. Upon an
accepted event, $\mathcal U$ advances the stage and updates the
memory contents from the current observation. This
bidirectional interaction lets the VLA remain Markovian while
the external program carries the long-horizon task state.

\subsection{Motor-Skill Learning in Weight Space}
\label{sec:weight_learning}

The motor component is a Markovian flow-matching policy
$\pi_\theta(\mathbf a_t\mid o_t,\ell)$ initialized from a
pretrained VLA, such as $\pi_{0.5}$ \citep{pi05}. Let
$\boldsymbol\epsilon\sim\mathcal N(0,I)$ denote Gaussian noise
and $\tau$ a flow timestep sampled uniformly from $[0,1]$. Following the
original flow-matching objective, we interpolate between noise
and an expert action chunk as
\begin{equation}
\mathbf a^\tau
=
(1-\tau)\boldsymbol\epsilon+\tau\mathbf a,
\end{equation}
and fine-tune the velocity field by
\begin{equation}
\theta^\star
=
\arg\min_\theta
\mathbb E_{\substack{
(o,\mathbf a,\ell)\sim\mathcal D,\\
\tau,\boldsymbol\epsilon}}
\left[
\left\|
v_\theta(\mathbf a^\tau,\tau\mid o,\ell)
-
(\mathbf a-\boldsymbol\epsilon)
\right\|_2^2
\right].
\label{eq:flow_matching}
\end{equation}
This stage learns the contact-rich motor behaviors demonstrated
in $\mathcal D$. Once fine-tuning is complete, $\theta^\star$
is fixed. Memory acquisition therefore does not modify the
policy weights or require additional expert actions.

\subsection{Memory-Strategy Learning in Code Space}
\label{sec:code_learning}

The memory component maintains an explicit symbolic state
\begin{equation}
s_t=
\bigl(\texttt{plan},\rho_t,\mathcal B_t,\mathcal N_t,\mathcal F_t\bigr),
\label{eq:memory}
\end{equation}
where $\rho_t$ denotes the active plan stage. The remaining
variables record task bindings, repeated-event counts, and
persistent interaction states, together realizing the estimate
of the task state $z_t$ in Eq.~\ref{eq:task_state}. Unlike a
learned history embedding,
$s_t$ is updated by executable rules and directly exposes the
task information used to select the next behavior.

The program $\mathcal P$ is developed iteratively by the coding agent. Given a candidate program $\mathcal P^{(n)}$, rollout $n$
produces a symbolic execution trace
\begin{equation}
\xi_n
=
\bigl(
(s_t,e_t,R_t)
\bigr)_{t=1}^{T_n},
\label{eq:trace}
\end{equation}
together with the stage-wise verdicts $\mathbf b_n$ provided by
the benchmark. The coding agent compares the stage transitions and
completion evidence recorded by PACE with $\mathbf b_n$ to
localize the failed stage. It then diagnoses whether the active
constraint reward is mis-specified or the corresponding
verification rule is overly strict or permissive, and revises the
executable program:
\begin{equation}
\mathcal P^{(n+1)}
=
\operatorname{Edit}
\left(
\mathcal P^{(n)},
\xi_n,
\mathbf b_n
\right).
\label{eq:program_update}
\end{equation}
The revision can modify the stage constraint reward or adjust the
event-verification and transition rules. The best-performing
program on development rollouts is retained as $\mathcal P^\star$.

This learning process updates software rather than neural-network
parameters. It uses neither gradients nor expert action labels,
allowing the coding agent to refine memory logic from execution
feedback. During evaluation, the coding agent applies the frozen program $\mathcal P^\star$ to update the within-episode memory and instantiate stage-specific rules and rewards, without rollout-level diagnosis or program revision.

\subsection{Memory-Conditioned Action Steering}
\label{sec:steering}

At control step $t$, the active stage $\rho_t$ and symbolic
memory $s_t$ select a differentiable constraint
$R_{\rho_t}(\mathbf a,o_t;s_t)$. For example, a reaching stage
may instantiate
\begin{equation}
R_{\mathrm{reach}}
=
-
\left\|
\operatorname{ee}(\mathbf a)
-
\Gamma\bigl(\operatorname{target}(s_t),o_t\bigr)
\right\|_2^2,
\label{eq:reach_constraint}
\end{equation}
where $\operatorname{ee}(\mathbf a)$ is the terminal
end-effector position of chunk $\mathbf a$ and
$\Gamma(\cdot)$ grounds a symbolic target into scene
keypoints using open-vocabulary detection, mask refinement, and
depth back-projection as in VLS \citep{vls}.

The selected constraint is injected into the flow-matching
velocity field:
\begin{equation}
\hat v
=
v_{\theta^\star}
(\mathbf a^\tau,\tau\mid o_t,\ell)
+
\lambda_t
\nabla_{\mathbf a^\tau}
R_{\rho_t}(\mathbf a^\tau,o_t;s_t),
\label{eq:guidance}
\end{equation}
where the constraint is evaluated on the intermediate chunk
following VLS \citep{vls}, and its gradient coincides with the
clean-action gradient as $\tau\to1$.
Integrating $\hat v$ steers the denoising process toward an action
mode consistent with the current memory state. The symbolic
program determines which behavior should be executed, while the
VLA retains responsibility for realizing the corresponding motor
skill.

We further attenuate guidance as the policy approaches the
interaction region, with the schedule
\begin{equation}
\lambda_t
=
\frac{\lambda_0}
{1+\exp\!\left(\beta(p_t-p_{\mathrm{mid}})\right)},
\label{eq:adaptive_guidance}
\end{equation}
where $\lambda_0$ is the base scale,
$p_t=\operatorname{clip}\bigl(1-c_t/c_{t_\rho},0,1\bigr)$ is
the normalized reduction of the constraint residual
$c_t=-R_{\rho_t}(\mathbf a_t,o_t;s_t)$ since stage entry at
step $t_\rho$, $p_{\mathrm{mid}}$ is the midpoint at which the
scale halves, and $\beta$ controls the decay rate. This schedule lets memory
guide global behavior selection while allowing the learned motor
policy to dominate contact-rich execution.

\subsection{Multimodal Progress Verification and Memory Update}
\label{sec:verification}

Reliable memory updates require determining whether the current
stage has actually been completed. We refer to this mechanism as
\emph{Proprioception-And-Completion-driven stagE-switching}
(PACE). PACE infers completion from two
complementary evidence streams available during robot execution.
Proprioceptive evidence detects action-dependent motion patterns
and preserves persistent events through symbolic latches. Visual
evidence is provided by Qwen3-VL-8B, which judges completion from
the active stage and up to five recent RGB frames. The multi-frame
context helps distinguish an ongoing action from a completed one.

PACE issues completion queries periodically; let $r(t)$ index
the most recent query before step $t$. At query $r$, the two
evidence streams produce binary judgments
$e_{\mathrm{prop}}^{(r)}$ and $e_{\mathrm{vlm}}^{(r)}$,
combined as
\begin{equation}
d_r
=
e_{\mathrm{prop}}^{(r)}
\lor
e_{\mathrm{vlm}}^{(r)}.
\label{eq:completion_judgment}
\end{equation}
To suppress isolated errors, the stage-progress event fires only
when completion is supported by at least $k$ of the $w$ most
recent queries:
\begin{equation}
e_t
=
\mathbb I
\Biggl[
\sum_{i=r(t)-w+1}^{r(t)}d_i\ge k
\Biggr],
\label{eq:completion_vote}
\end{equation}
which instantiates the verification rule $\mathcal V$ of
Eq.~\ref{eq:overview_event} over the window $\mathcal W_t$
covered by these $w$ queries; the vote is reset upon stage
advancement. We use $w=5$ and $k=3$ in all experiments.
Once an event is accepted, $\mathcal U$ updates the corresponding
bindings, counters, and persistent states before activating the
constraint for the next stage. This closes the loop between
memory-conditioned steering and physical execution without
requiring privileged environment states or separately trained task-specific completion models.

\section{Experiments}

\subsection{Experimental Setup}
\label{sec:experimental_setup}

\textbf{Tasks and metrics.}
We evaluate \method{} on a corrected 12-task protocol from
RoboMemArena~\citep{robomemarena}, comprising 160 episodes across
transferring, counting, sequential execution, and occlusion.
We correct success checks that can produce false positives and retain
tasks where $\pi_{0.5}$ has the required motor skills but still exhibits
failures caused by missing history. Tasks solvable through observation
shortcuts or dominated by motor-skill failures are excluded. This
selection uses only protocol checks and $\pi_{0.5}$ behavior, without
\method{} results. All methods are evaluated on the same corrected
tasks and episodes. Each transferring and counting task uses 10
episodes, each occlusion task uses 15, and each sequence task uses 20.
We additionally evaluate three memory-dependent tasks on an SO-101
robot. We report Task Success Rate (TSR) and Cumulative Success Rate
(CSR), which measure full-task success and stage-level progress,
respectively. Exploration experience is frozen during evaluation.

\smallskip
\noindent
\textbf{Baselines.}
We compare against the reactive $\pi_{0.5}$ policy and PrediMem
\citep{robomemarena}, the memory-augmented baseline introduced
with RoboMemArena. PrediMem combines a high-level VLM planner
with recent-frame and keyframe memory, while \method{} steers a
Markovian VLA using memory strategies learned in executable code.
PrediMem is reevaluated on our 12-task protocol rather than using
its published 26-task results. On the real robot, $\pi_{0.5}$ and
\method{} share the same demonstrations and fine-tuned policy
weights, isolating the effect of executable memory.

\smallskip
\noindent
\textbf{Implementation details.}
Simulation uses the RoboMemArena-trained $\pi_{0.5}$
 checkpoint through OpenPI, while the
real-robot policy is fine-tuned on SO-101 demonstrations through
LeRobot. Policy weights remain fixed during heuristic development
and evaluation. The coding agent uses an Opus 4.8 model to update
experience from PACE traces and benchmark stage verdicts during
development. During evaluation, it applies the frozen program
$\mathcal P^\star$ to instantiate context-dependent constraints;
rollout summarization, experience revision, and updates to
$\mathcal P^\star$ are disabled.
Qwen3-VL-8B is used only for multi-frame stage-completion verification.

\subsection{Quantitative Results}
\label{sec:quantitative_results}

\begin{table*}[t]
\centering
\small
\begin{tabular*}{\textwidth}{@{\extracolsep{\fill}}p{0.40\textwidth}cccccc@{}}
\toprule
& \multicolumn{3}{c}{CSR} & \multicolumn{3}{c}{TSR} \\
\cmidrule(lr){2-4}\cmidrule(lr){5-7}
Scenario & $\pi_{0.5}$ & PrediMem & \method{} & $\pi_{0.5}$ & PrediMem & \method{} \\
\midrule
\multicolumn{7}{l}{\textbf{Transferring}} \\
Pudding + butter $\rightarrow$ cabinet 2 & 90.0 & \textbf{100.0} & \textbf{100.0} & 80.0 & \textbf{100.0} & \textbf{100.0} \\
Butter + cheese $\rightarrow$ plate 2 & 50.0 & \textbf{80.0} & \textbf{80.0} & 50.0 & 70.0 & \textbf{80.0} \\
Pudding + cheese $\rightarrow$ plate 2 & 70.0 & 80.0 & \textbf{100.0} & 70.0 & 60.0 & \textbf{100.0} \\
\textit{Category average} & 70.0 & 86.7 & \textbf{93.3} & 66.7 & 76.7 & \textbf{93.3} \\
\addlinespace
\multicolumn{7}{l}{\textbf{Counting}} \\
Tomato sauce $\times2$ on cookies $\rightarrow$ drainer & 50.0 & \textbf{86.7} & 63.0 & \textbf{60.0} & \textbf{60.0} & \textbf{60.0} \\
Pudding + pour $\times2$ $\rightarrow$ drainer & 35.0 & \textbf{57.5} & 50.0 & 0.0 & 20.0 & \textbf{40.0} \\
Butter + pour $\times2$ $\rightarrow$ drainer & 32.5 & \textbf{72.5} & 68.0 & 0.0 & 30.0 & \textbf{50.0} \\
\textit{Category average} & 39.2 & \textbf{72.2} & 60.3 & 20.0 & 36.7 & \textbf{50.0} \\
\addlinespace
\multicolumn{7}{l}{\textbf{Sequence}} \\
Butter + popcorn $\rightarrow$ basket & 60.0 & 72.5 & \textbf{80.0} & 40.0 & 50.0 & \textbf{65.0} \\
Cream + pudding $\rightarrow$ basket & 45.0 & \textbf{67.5} & \textbf{67.5} & 40.0 & 50.0 & \textbf{60.0} \\
\textit{Category average} & 52.5 & 70.0 & \textbf{73.8} & 40.0 & 50.0 & \textbf{62.5} \\
\addlinespace
\multicolumn{7}{l}{\textbf{Occlusion}} \\
Cookies + chocolate $\rightarrow$ microwave & \textbf{55.0} & 38.3 & 42.0 & \textbf{46.7} & 26.7 & 28.0 \\
Butter + chocolate $\rightarrow$ microwave & 58.3 & 36.7 & \textbf{63.3} & 46.7 & 40.0 & \textbf{80.0} \\
Cream + popcorn $\rightarrow$ microwave & 50.0 & 50.0 & \textbf{57.0} & 46.7 & 46.7 & \textbf{60.0} \\
Cookies + popcorn $\rightarrow$ microwave & 38.3 & 28.3 & \textbf{40.0} & \textbf{20.0} & 13.3 & \textbf{20.0} \\
\textit{Category average} & 50.4 & 38.3 & \textbf{50.6} & 40.0 & 31.7 & \textbf{47.0} \\
\midrule
\textbf{Overall} & 52.5 & 61.7 & \textbf{66.2} & 41.3 & 45.6 & \textbf{60.1} \\
\bottomrule
\end{tabular*}
\caption{Comparison of the fine-tuned $\pi_{0.5}$ policy, PrediMem, and \method{} on RoboMemArena. Rows group tasks by memory capability, while columns report Cumulative Success Rate (CSR), which measures average subgoal completion, and Task Success Rate (TSR), which measures full-task completion. Category-average rows summarize each task category, and the final row reports overall performance across all 160 evaluation episodes. \method{} executes the best-performing heuristic selected during exploration and freezes it throughout evaluation, with code editing and reflection disabled. Bold denotes the best result.}
\label{tab:robomemarena_results}
\end{table*}

\begin{table}[t]
\centering
\footnotesize
\renewcommand{\arraystretch}{1.15}
\begin{tabular}{@{}p{0.40\columnwidth}cc@{}}
\toprule
Task & $\pi_{0.5}$ & \method{} \\
\midrule
{\raggedright Observe-and-pick-up\par} &
2/10 & \textbf{7/10} \\
\addlinespace[2pt]
{\raggedright Number-Guided Button Pressing\par} &
7/15 & \textbf{11/15} \\
\addlinespace[2pt]
{\raggedright Put-Back Block\par} &
0/10 & \textbf{2/10} \\
\bottomrule
\end{tabular}
\caption{Real-world task success on the SO-101. Columns compare the fine-tuned $\pi_{0.5}$ policy with \method{}. \method{} executes the heuristic selected during exploration and keeps it fixed throughout evaluation.}
\label{tab:real_world_results}
\end{table}

\textbf{Simulation experiments.}
Table~\ref{tab:robomemarena_results} presents the task-level,
category-averaged, and overall CSR and TSR of $\pi_{0.5}$,
PrediMem, and \method{} on 12 RoboMemArena tasks. All methods
are evaluated on the same tasks and scene configurations across
160 evaluation episodes.
The results yield four main findings. i) Under identical policy weights,
\method{} improves CSR from 52.5\% to 66.2\% and TSR from
41.3\% to 60.1\% over $\pi_{0.5}$, showing that executable
memory and steering compensate for the lack of built-in memory
in a Markovian VLA. ii) \method{} exceeds PrediMem by 14.5 TSR
points, despite a smaller 4.5-point CSR gain. This result indicates
that \method{} more effectively converts intermediate progress into
full-task success, supporting our hybrid design that couples
weight-space motor learning with code-space memory-strategy learning.
iii) The largest TSR gains over $\pi_{0.5}$ occur on counting
(30.0 points), transferring (26.6 points), and sequence tasks
(22.5 points), consistent with the benefits of explicit counters,
object--location associations, and stage state. On counting tasks, \method{} achieves
higher TSR than PrediMem (50.0\% vs.\ 36.7\%) despite lower CSR
(60.3\% vs.\ 72.2\%). In our experiments, we found that these
failures were concentrated in early execution stages, where they
may stem from limitations of the frozen VLA or occasional guidance
interference. Once these stages succeed,
explicit counting and verified transitions make full-task
completion more reliable.
iv) Occlusion is the hardest category for memory injection:
PrediMem falls below the reactive $\pi_{0.5}$ (38.3\% vs.\
50.4\% CSR), suggesting that an unreliable memory can actively
mislead execution when the target is hidden. \method{} matches
$\pi_{0.5}$ on CSR (50.6\%) while improving TSR by 7.0 points,
indicating that verified stage transitions limit the damage of
wrong bindings, though visual completion evidence remains least
reliable under occlusion.

\smallskip
\noindent
\textbf{Real-world experiments.}
Table~\ref{tab:real_world_results} reports the task success rates
of $\pi_{0.5}$ and \method{} on three memory-dependent tasks
using the SO-101 robot, with 35 trials per method in total.
The two methods share the same demonstrations, observations, and
fine-tuned policy weights; only \method{} maintains symbolic
memory and applies memory-conditioned steering. Three findings
emerge. i) Overall TSR rises from 25.7\% to 57.1\%, validating
executable memory and steering on hardware. ii) Per-task gains of
50.0, 26.7, and 20.0 points show that one symbolic state supports
cue retention, counting, and source--location associations.
iii) Gains in the first two tasks support history-conditioned
selection and verified transitions; Put-Back remains challenging
because variations in ambient illumination reduce the robustness
of the underlying visuomotor policy.

\smallskip
\noindent
\textbf{Demonstration cost and inspectability.}
The two properties that motivate our design are also visible in
these experiments. First, all results above are obtained with a
single fine-tuned checkpoint per platform: acquiring memory never
updates policy weights, and a new memory configuration---a
different target color or a different required count---is realized
by editing the code-space program instead of by collecting
demonstrations that cover it. The demonstration budget therefore
scales with the number of reusable motor skills rather than with
the number of history-dependent task configurations. Second, the
symbolic state is readable at every control step, so each failed episode can be localized to memory update, motor execution, or event verification. We use this property below to
analyze the remaining failures.

\subsection{Qualitative Results}
\label{sec:qualitative_results}

\begin{figure}[!t]
\centering
\includegraphics[width=\columnwidth]{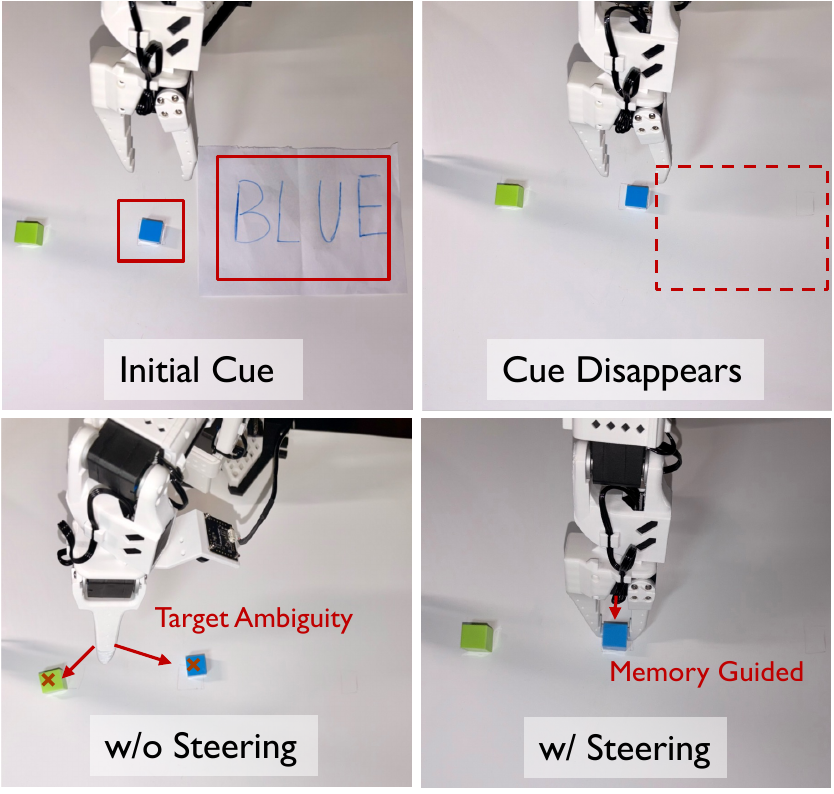}
\caption{\textbf{Memory-guided execution on the SO-101
(Observe-and-Pick-Up).} \emph{Top:} the target color is written on a
card that is visible only at the start of the episode (\emph{Initial
Cue}) and is then removed from the workspace (\emph{Cue Disappears}),
so the current observation no longer identifies which block to pick.
\emph{Bottom:} without steering, the reactive $\pi_{0.5}$ policy no
longer has access to the cue and reaches indifferently toward either
candidate block (\emph{Target Ambiguity}). \method{} retains the cue
in executable memory and instantiates the corresponding constraint,
which steers the same policy weights to the blue block
(\emph{Memory Guided}).}
\label{fig:qualitative_results}
\end{figure}

\textbf{Visualization of memory-guided execution.}
Figure~\ref{fig:qualitative_results} illustrates how \method{}
uses executable memory to resolve history-dependent action
ambiguity. The top row shows the transient task cue and the
current observation after the cue has been removed; the current observation alone provides no evidence about which block is the target.
The bottom row contrasts the two policies under the same
observation. Although the reactive $\pi_{0.5}$ policy can execute
the required motor skill, it does not select a history-consistent
target because the relevant history is absent from its input.
\method{} instead keeps the cue as symbolic state, translates it
into a memory-conditioned constraint, and verifies stage
completion with proprioceptive and multi-frame visual evidence
before activating the next constraint. We see that \method{}
converts information from an earlier observation into an explicit
control decision, enabling the same Markovian VLA to produce the
history-consistent action.

\begin{figure}[!t]
\centering
\includegraphics[width=\columnwidth]{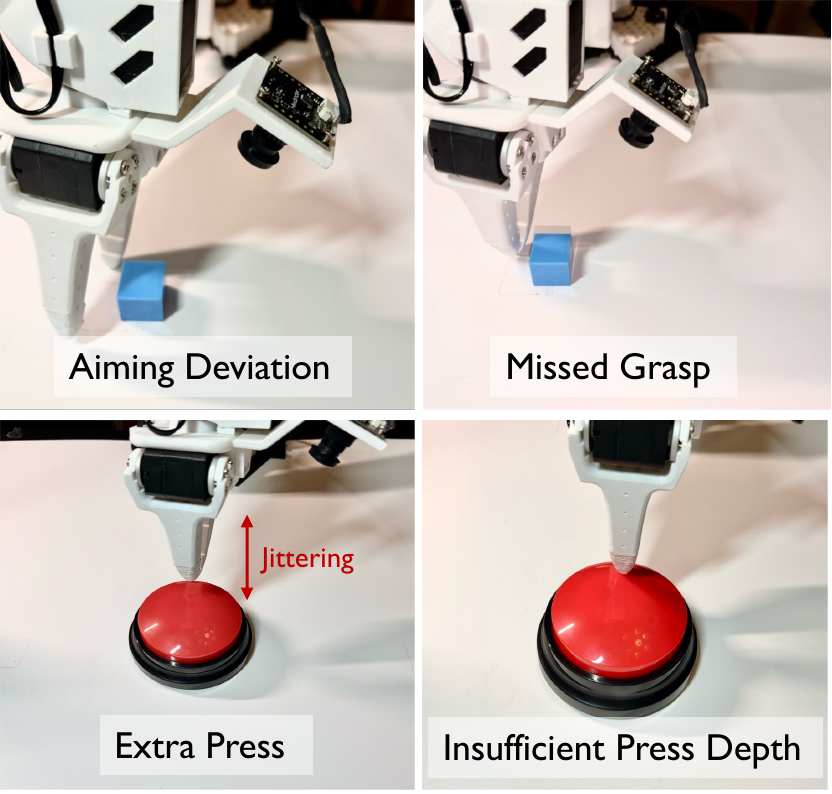}
\caption{\textbf{Interpretable failure modes on the SO-101.}
Across these examples, the symbolic memory and activated constraint
are correct before execution. The failures instead arise from motor
execution or stage-completion verification.
\emph{Top:} a lateral deviation at the pre-grasp pose
(\emph{Aiming Deviation}) makes the gripper close beside the
remembered target (\emph{Missed Grasp}). \emph{Bottom:} on the
button task, vertical jitter near the contact point causes an
unintended second press (\emph{Extra Press}), whereas an
insufficient press depth (\emph{Insufficient
Press Depth}) is not recognized as completion by PACE.}
\label{fig:qualitative_failure}
\end{figure}

\begin{table}[t]
\centering
\footnotesize
\setlength{\tabcolsep}{3.5pt}
\begin{tabular}{@{}lcccc@{}}
\toprule
Variant & Learn & PACE evidence & CSR & TSR \\
\midrule
One-shot $\mathcal P^{(0)}$ & -- & Vis.+Prop. & 63.3 & 53.3 \\
Vision-only PACE & \checkmark & Vis. & 55.8 & 50.0 \\
Proprio.-only PACE & \checkmark & Prop. & 63.0 & 63.3 \\
Full \method{} & \checkmark & Vis.+Prop. & \textbf{71.8} & \textbf{71.7} \\
\bottomrule
\end{tabular}
\caption{Ablations on a six-task subset under shared policy weights and evaluation configurations. We ablate heuristic refinement and the visual and proprioceptive evidence used by PACE. CSR and TSR are percentages; the dash marks the variant that does not use rollout-driven heuristic learning.}
\label{tab:ablation}
\end{table}

\smallskip
\noindent
\textbf{Visualization of failure modes.}
Figure~\ref{fig:qualitative_failure} presents representative
failures in which the symbolic state and activated constraint are
initially correct, but motor execution or stage-completion verification
fails. The top row shows a grasp failure: the remembered
target and the steering direction are correct, but an aiming
deviation at the pre-grasp pose causes the pretrained VLA to close
the gripper beside the block. The bottom row shows two
counting failures on the button task, where execution jitter
produces a spurious extra press, while a shallow press is not
recognized as complete by PACE. These failures arise from motor
execution and event verification rather than from long-horizon
memory corruption.

\subsection{Ablation Studies}

Table~\ref{tab:ablation} evaluates rollout-driven heuristic learning and multimodal stage-completion verification on a six-task subset spanning all four task categories (one to two tasks per category). All variants use the same fine-tuned $\pi_{0.5}$ policy, observations, and evaluation episodes. The ablation yields two findings. i) Replacing the one-shot program $\mathcal P^{(0)}$ with the rollout-refined program $\mathcal P^{\star}$ improves CSR and TSR by 8.5 and 18.4 points, validating rollout experience as effective feedback for learning memory-management heuristics. ii) Relying on either completion signal alone degrades performance: visual-only verification reduces CSR and TSR by 16.0 and 21.7 points, while proprioception-only verification reduces them by 8.8 and 8.4 points. Either signal alone can misjudge stage transitions; their complementary evidence makes these transitions more reliable.

\section{Conclusion}

We propose \method{}, a hybrid learning framework for
memory-dependent robot manipulation. The core idea is to learn motor
skills in weight space through gradient-based imitation learning,
while acquiring memory-management strategies in code space
through heuristic learning. The resulting executable memory
selects stage-specific constraints to steer a Markovian VLA, while
multimodal progress verification updates the memory from physical
execution. Experiments on RoboMemArena and real-world manipulation
tasks show that \method{} substantially improves stage completion
and task success without requiring expert demonstrations to cover
every history-dependent task configuration.

\smallskip
\noindent
\textbf{Limitations and future work.}
The current work focuses on tasks whose relevant history can be
represented by a compact symbolic state. Two further limitations
remain. Task success remains limited by both motor competence and stage-completion verification. As the failure analysis shows, a correct memory
state and an appropriate constraint do not guarantee successful
contact-rich execution. In addition, the periodic VLM queries
and coding-agent calls add inference latency and cost per
episode.
In future work, we plan
to extend the same hybrid learning principle to hierarchical and
multimodal memory representations, enabling more open-ended
instructions and longer-horizon manipulation.

\bibliography{aaai2027}


\end{document}